\documentclass[]{spie}  

\usepackage{amsmath,amsfonts,amssymb}
\usepackage{graphicx}
\usepackage[colorlinks=true, allcolors=blue]{hyperref}
\usepackage{wrapfig}
\usepackage{subfigure}

\title{Learning Numerical Observers using Unsupervised Domain Adaptation}
\vspace{-.5cm}
\author[1]{Shenghua He}
\author[2]{Weimin Zhou}
\author[3,4]{Hua Li}
\author[4]{Mark A. Anastasio}
\affil[1]{Department of Computer Science and Engineering,}
\affil[2]{Department of Electrical and Systems and Engineering,\break Washington University in St$.\,$Louis, St$.\,$Louis, MO 63130, USA}
\affil[3]{Carle Cancer Center, Carle Foundation Hospital, Urbana, IL, 61801}
\affil[4]{Department of Bioengineering, \break University of Illinois at Urbana-Champaign, Urbana, IL 61801, USA}

\authorinfo{Further author information: (Send correspondence to Mark A. Anastasio.)\\E-mail: maa@illinois.edu, Telephone: 1 314 935 3637}

\begin{document} 
\maketitle
\vspace{-.2cm}
\begin{abstract}
Medical imaging systems are commonly assessed by use of objective image quality measures. Supervised deep learning methods have been investigated to implement numerical observers for task-based image quality assessment. 
However, labeling large amounts of experimental data to train deep neural networks is tedious, expensive, and prone to subjective errors.
Computer-simulated image data can potentially be employed to circumvent these issues; however, it is often difficult to computationally model complicated anatomical structures, noise sources, and the response of real-world imaging systems. Hence, simulated image data will generally possess physical and statistical differences from the experimental image data they seek to emulate.  Within the context of machine learning, these differences between the sets of two images is referred to as \emph{domain shift}. In this study, we propose and investigate the use of an adversarial domain adaptation method to mitigate the deleterious effects of domain shift between simulated and experimental image data for deep learning-based numerical observers (DL-NOs) that are trained on simulated images but applied to experimental ones. In the proposed method, a DL-NO will initially be trained on computer-simulated image data and subsequently adapted for use with experimental image data, without the need for any labeled experimental images. As a proof of concept, a binary signal detection task is considered.
The success of this strategy as a function of the degree of domain shift present between the simulated and experimental image data is investigated.


\end{abstract}
\vspace{-.2cm}
\section{Introduction}
Medical imaging systems are commonly assessed by use of objective measures of image quality that quantify the performance of an observer at specific tasks\cite{barrett2013foundations, zhou2018learning, zhou2019learningHO, zhou2019learning, zhou2019approximating,zhou2019learningSOM}. Supervised deep learning methods have been actively investigated to learn and implement numerical observers for task-based image quality assessment. For example, Zhou \textit{et. al.} have proposed an Ideal Observer approximation methodology for binary signal detection tasks by use of convolutional neural networks (CNNs)~\cite{zhou2019approximating}. These  supervised deep learning-based methods require a large amount of labeled data for training. 
However, in practice, labeling a large of experimental data is tedious, expensive, and prone to subjective errors. 

In contrast, labeled computer-simulated image data
 can be relatively convenient to generate. If the simulated data are realistic enough, it is potentially feasible to train a deep learning-based numerical observer (DL-NO) with a large amount of
 simulated data and then directly apply it to experimental data.
 However, it is often difficult to computationally model complicated
 anatomical structures and the response of real-world imaging systems and therefore
simulated image data will generally possess physical and statistical differences from the
 experimental image data they seek to emulate.  This results in a so-called
\emph{domain shift} between the two sets of images\cite{gre2009, gopalan2011domain, ganin2014unsupervised}.
This domain shift can significantly degrade the performance of a 
DL-NO that is trained on simulated images but applied to experimental ones.
Recently, domain adaptation methods that aim at mitigating the effect of domain shifts have been applied to several computer vision tasks including image classification~\cite{tzeng2017adversarial, kang2019contrastive, lee2019sliced}, image segmentation~\cite{javanmardi2018domain,he2018convolutional,dou2018unsupervised} and cell counting~\cite{he2019automatic,he2019}.

In this study, we propose and investigate the use of an adversarial domain adaptation method to mitigate the deliterious effects of domain shift between simulated and experimental image data for DL-NOs that
are trained on simulated images but applied to experimental ones.
The employed domain adaption methodology will not require labelled experimental images.
 As a proof of concept, a convolutional neural network (CNN) is employed as the NO and  a
 binary signal detection task is considered. Through computer-simulation studies, the success of this strategy as a function of the degree of domain shift present between the simulated and experimental image data is investigated.
\vspace{-.2cm}
\section{Methods}
\subsection{Framework of the Proposed Method}
\begin{figure}
	\begin{center}
	    \includegraphics[width=0.9\textwidth]{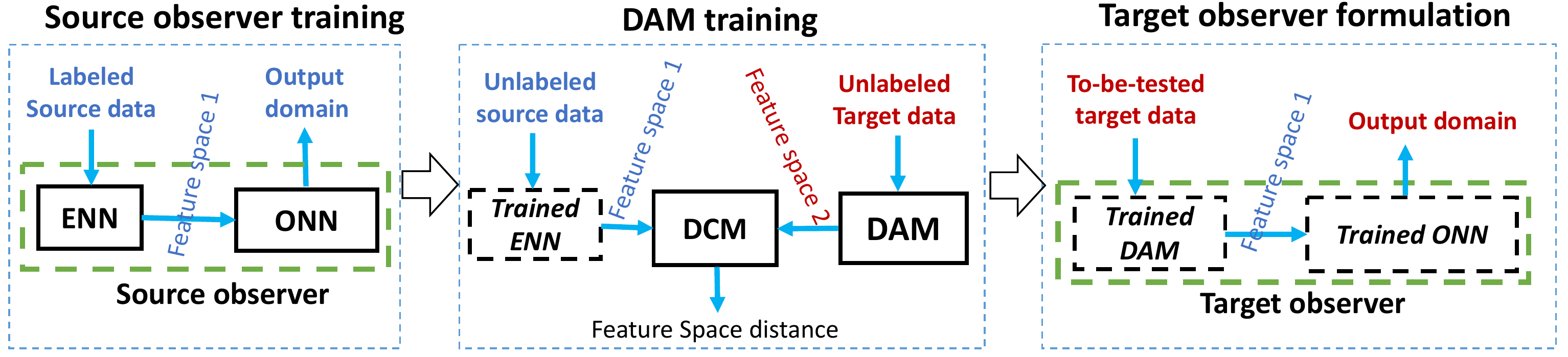}
	\end{center}
\vspace{-.3cm}
\caption{Overview of the proposed method.}
\label{fig:framework}
\vspace{-.5cm}
\end{figure}

The framework of the proposed method consists of three stages: source observer training, domain adaptation model (DAM) training, and target observer formulation, as shown in Figure~\ref{fig:framework}.

In the stage of source observer training, a large amount of labeled computer-simulated data (source data) are automatically generated, and then employed to train a DL-NO operating in the source domain (source observer). As shown in Figure~\ref{fig:framework}, the source observer contains an encoder neural network (ENN) and observation neural network (ONN). The ENN encodes a source data into a feature space that highly represents the source domain data. The ONN maps the encoded features to the desired output (e.g. test statistics for signal detection tasks).

In the DAM training stage, a deep neural network-based domain adaptation model (DAM) is trained by use of labeled simulated data (source data) and unlabeled experimental data (target data). The trained DAM will be employed to adapt the trained source observer to the target data domain (target domain). This task is achieved by mapping the target data to a feature space that is close to the feature space the trained ENN maps to. The DAM is trained by minimizing the distance between the feature space of the source domain and that of the target domain. A neural network-based domain critic model (DCM) is built up for measuring the distance between the two feature spaces. The DAM and DCM are iteratively trained via an adversarial learning approach introduced in the literature~\cite{arjovsky2017wasserstein}.

In the stage of target observer formulation, the trained DAM and the trained ONN (the second part of the source observer) are integrated to formulate a target numerical observer that can operate on experimental data for image quality assessment tasks.


\vspace{-.2cm}
\subsection{Example of the Proposed Method}
\label{sec:application}
\vspace{-.1cm}
In this study, the method for learning a CNN-based numerical observer for a binary signal detection task proposed by Zhou \textit{et. al.}~\cite{zhou2019approximating} is employed as an example to demonstrate the stages of the proposed method and its performance.

\subsubsection{Binary signal detection tasks}
The considered task is a binary signal detection task in which the goal is to classify an image $\mathbf{g} \in \mathbb{R}^{M\times 1}$ into either a signal-absent hypothesis ($H_0$) or a signal-present hypothesis ($H_1$). The imaging process under these two hypothesises can be represented as: 
\begin{equation}
\begin{aligned}
&H_0: \mathbf{g} = \mathbf{b}+\mathbf{n},\\
&H_1: \mathbf{g} = \mathbf{b} +\mathbf{s} +\mathbf{n},
\end{aligned}
\end{equation}
where $\mathbf{b}\in \mathbb{R}^{M\times 1}$ and $\mathbf{s} \in \mathbb{R}^{M\times 1}$ denote the background and signal in the image domain, respectively, and $\mathbf{n} \in \mathbb{R}^{M\times 1}$ is the measurement noise. Here $M$ is the total number of pixels in an image. 

A numerical observer computes a scalar test statistic $t$ for this binary signal detection task. A decision is made in favor of hypothesis $H_1$ if $t$ is greater than some threshold; otherwise $H_0$ is selected.

\subsubsection{Source observer training}
\label{subsec:source}
The goal of this stage is to train a CNN-based source observer by use of a large amount of simulated images
that represent the source domain.
This is depicted in the left block of Figure~\ref{fig:framework}.

In the stage of source observer training, the CNN-based source observer is trained. Figure~\ref{fig:architecture} shows the network architectures of the ENN and ONN. The ENN contains a chain of $7$ convolutional layer-Leaky ReLu layer (CONV-LeReLu) blocks and a max pooling layer (Max-Pool). The ONN has a fully connected layer-Leaky LeReLU (FC-LeReLU) block, followed by a sigmoid function in the last layer. The ENN encodes a simulated image into a low-dimensional but highly-representative feature space, while the ONN works as a classifier that computes a probability of the input simulated image belonging to hypothesis $H_1$ by use of the encoded features.

Let the CNN-based source observer be parameterized by a set of parameters $\Theta$. The output of the source observer can be represented by $p(H_1|\mathbf{g},\Theta)$.  
Let $y\in\{0,1\}$ denote the image label, where $y=0$ and $y=1$ correspond to the hypothesis $H_0$ and $H_1$, respectively. Given a set of $N_d$ independent labeled source images, $D = \{(\mathbf{g_i},y_i)\}_{i=1}^{N_d}$, $\Theta$ is determined by minimizing an average cross-entropy loss function, $l(\Theta|D)$, defined as:
\begin{equation}
l(\Theta|D) = -\sum_{i=1}^{N_d} y_i\log (p(y_i|\mathbf{g_i},\Theta)) + (1-y_i) (1-p(y_i|\mathbf{g_i},\Theta)),
\end{equation}
where $\mathbf{g}_i \in \mathbb{R}^{M\times1}$ and $y_i \in \{0,1\}$ are the $i^{th}$ training image and the associated label. The loss function $l(\Theta|D)$ is numerically minimized by use of methods described in the literature~\cite{zhou2019approximating}. The trained ENN (the first part of the source observer) will be employed for training the DAM as described next.

\subsubsection{Domain adaptation model (DAM) training}
\vspace{-.1cm}
The goal of DAM training stage is to train a DAM by use of a set of unlabeled source images (simulated images in the source domain), a set of unlabeled target images (experimental images in the target domain), and the trained ENN. The trained DAM will be employed to map target images to a feature space that has minimum domain shift with the feature space the trained ENN maps to.


\begin{figure}[!htb]
     \centering
     \includegraphics[width=0.55\textheight]{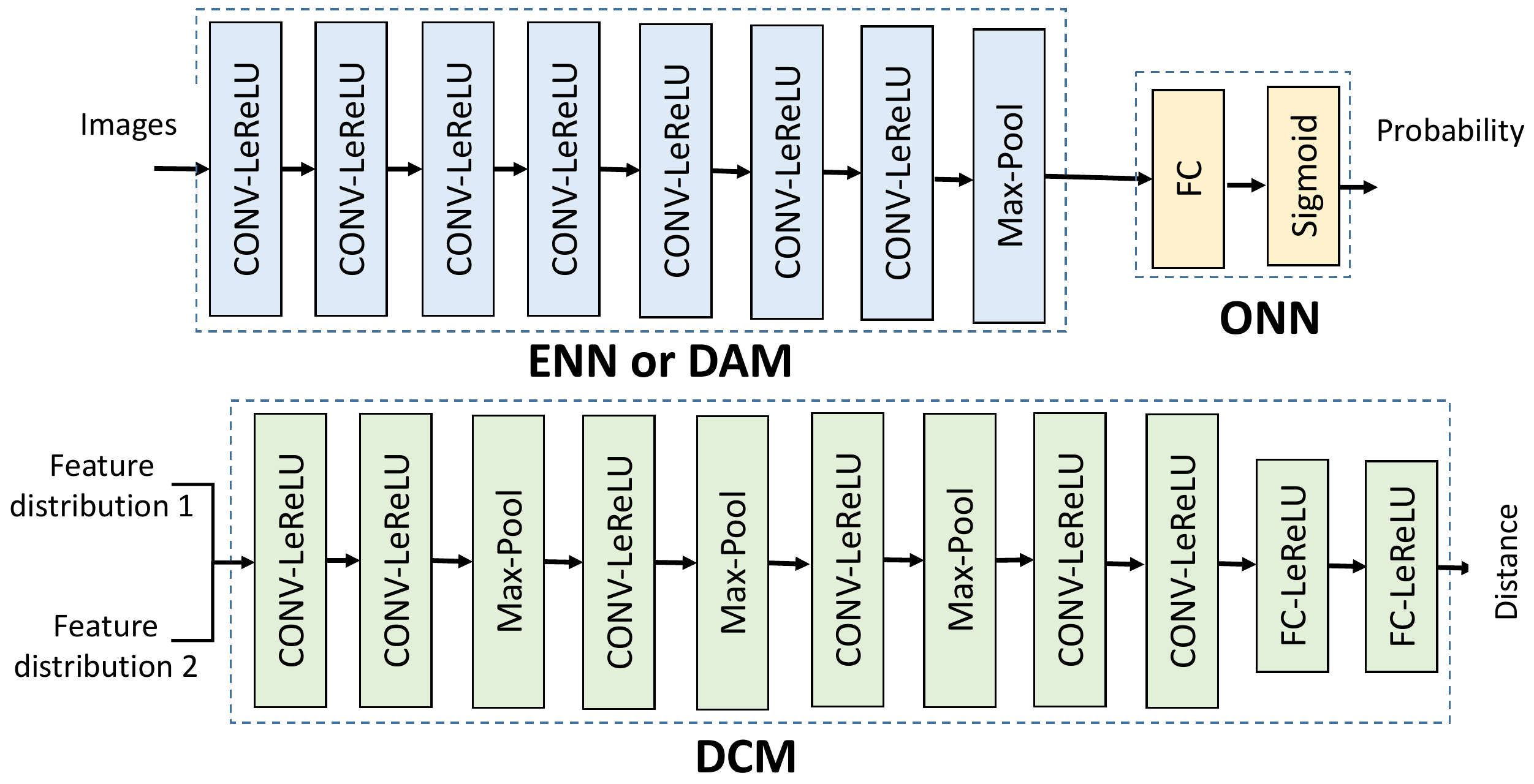}
     \caption{The network architectures of the ENN, ONN, DAM, and DCM. In this study, the network architectures of the ENN and DAM are the same.}\label{fig:architecture}
\end{figure}

The network architectures of the DAM and DCM are specified as two CNNs shown in Figure~\ref{fig:architecture}. In this study, the DAM has the same architecture as the ENN, considering both of them have similar function: mapping an image data into a low-dimensional feature space. The Wasserstein distance was employed to quantify the domain shift in the adversarial learning process for training the DAM and DCM. The trained CNN-based DAM will be employed as part of the target observer in the next stage.

\vspace{-.2cm}
\subsubsection{Target observer formulation}

In the final stage, the target numerical observer operating on experimental images is formulated by
combining the trained DAM (trained in the second stage) and the trained ONN (trained in the first stage), as shown in the right block of Figure~\ref{fig:framework}.
\subsection{Numerical Studies to Demonstrate the Performance of the Proposed Method}

In our proof-of-principle study, simulated images were employed to represent both the source
and target domain images. This permitted a controlled and systematic investigation of the proposed method.
The experimental (target domain) images were assumed to be produced by an idealized parallel-hole collimator system described by a point response function of the form~\cite{kupinski2003ideal}
\begin{equation}
h_m(\mathbf{r}) = \frac{h}{2\pi w^2} \exp \Big(-\frac{(\mathbf{r}-\mathbf{r}_m)^T(\mathbf{r}-\mathbf{r}_m)}{2w^2}\Big),
\end{equation}
where the system height $h=50$. Five sets of target domain images were produced with $5$ different system blurs $w = 2.0 $, $w = 3.0 $, $w = 4.0 $, $w = 5.0 $ and $w = 6.0 $, respectively. The simulated (source domain) images were produced by the same imaging model but with an incorrect value of the system height $h=40$ and that of the system blur $w = 0.5$. It can be expected that different sets of target domain images have different levels of domain shifts with the source domain images. These data were employed to investigate how the degree of domain shift impacts the performance of the proposed method.

A binary signal-known-exactly and background-known-statistically (SKE/BKS)  detection task
was considered.
For both the source and target domain images,  the signal function, $f_s(\mathbf{r})$, was described by a 2D symmetric Gaussian function:
\begin{equation}
f_s(\mathbf{r}) = A \exp \Big(-\frac{(\mathbf{r}-\mathbf{r}_c)^T(\mathbf{r}-\mathbf{r}_c)}{2w_s^2}\Big),
\end{equation}
where $A=0.2$ is the amplitude, $\mathbf{r}_c = [32,32]^T$ is the coordinate of the signal location, and $w_s = 3$ is the width of the signal. The background described by a stochastic lumpy object model:
\begin{equation}
f_b(\mathbf{r}) = \sum_{n=1}^{N_b} l(\mathbf{r}-\mathbf{r}_n|a,s),
\end{equation}
where $N_b$ is the number of lumps that is sampled from a Poisson distribution: $N_b\sim \mathcal{P}(\overline{N})$, where $\mathcal{P}(\overline{N})$ denotes a Poisson distribution with the mean $\overline{N}$ that was set to $5$, and the $ l(\mathbf{r}-\mathbf{r}_n|a,s)$ is the lumpy function modeled by a 2D Gaussian function with amplitude $a$ and width $s$:
\begin{equation}
l(\mathbf{r}-\mathbf{r}_n|a,s) = a \exp \Big(-\frac{(\mathbf{r}-\mathbf{r}_n)^T(\mathbf{r}-\mathbf{r}_n)}{2s^2}\Big).
\end{equation}
Here, $a$ was set to $1$, $s$ was set to $7$, and $\mathbf{r}_n$ is the location of the $n^{th}$ lumpy that was sampled from a uniform distribution over the field of view.

The signal image $\mathbf{s}$, the background image $\mathbf{b}$, and measurement noise $\mathbf{n}$ were then generated as described below. 
The $m^{th}$ pixel of a signal image $\mathbf{s}$ and background image $\mathbf{b}$ were computed as:
\begin{equation}
 s_m= \frac{Ah\omega_s^2} {(\omega^2+\omega_s^2)} \exp \Big(-\frac{(\mathbf{r}_m-\mathbf{r}_c)^T(\mathbf{r}_m-\mathbf{r}_c)}{2(\omega^2+\omega_s^2)}\Big),
\end{equation}
and 
\begin{equation}
b_m = \frac{ahs^2}{\omega^2+s^2} \sum_{n=1}^{N_b}\exp \Big(-\frac{(\mathbf{r}_m-\mathbf{r}_n)^T(\mathbf{r}_m-\mathbf{r}_n)}{2(\omega^2+s^2)}\Big).
\end{equation}
The measurement noise was described by independent and identically distributed Gaussian random variables that models electronic noise: $n_m\sim \mathcal{N}(0,\delta^2)$, where $\mathcal{N}(0,\delta^2)$ denotes a Gaussian distribution with the mean $0$ and the standard deviation $\delta$, which was set to $10$ in this study. 

\begin{figure}[!htb]
    \centering
     \includegraphics[width=0.7\textheight]{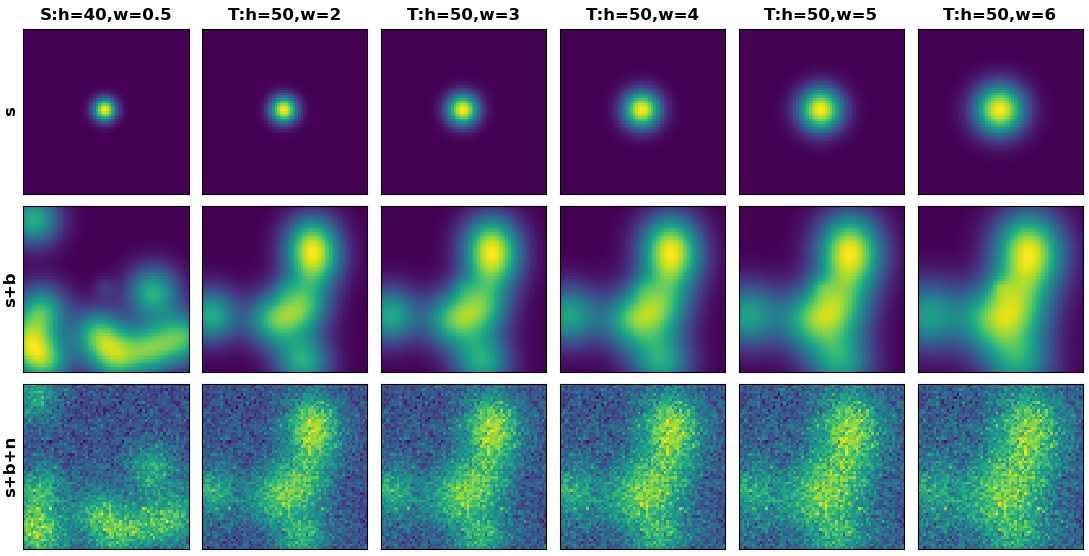}
     \caption{Example images in the source and target domains. The column of S:h=40, w=0.5 represents a signal image, background image, and noisy signal-present example in the source domain. Each of the remaining columns represents a signal image, background image, and noisy signal-present example in one of the 5 target domains, respectively, which were produced by a system model with $h=50$ and different system blurs.}
     \label{fig:data}
\end{figure}

The image sizes of both source and target images were $64\times 64$ pixels (i.e. $M= 4096$). 
Examples of signal-present images in the source and 5 target domains are shown Figure~\ref{fig:data}. 
Here, $100,200$ pairs of signal-present and signal-absent source images with labels were generated. Out of these image pairs, $100,000$ were employed as the training data to train the CNN-based source observer (shown in Figure~\ref{fig:architecture}), and $200$ were used as the related validation set for validating the source observer.
During the source observer training, the values of parameters that result in the highest AUC performance of the source observer evaluated on the validation set were selected for the ENN and ONN.

Additionally, in each of the 5 target image sets, $5,000$ pairs of signal-present and signal-absent target images without labels and $200$ with labels were generated. Five DAMs were trained, and each of them was associated with one of the 5 target image sets and used for adapting the trained source observer to the corresponding set of target images. In each DAM training, the unlabeled $100,000$ source image pairs (generated in the previous stage) and the $5,000$ unlabeled target image pairs from the corresponding target domain were employed for the adversarial learning of the CNN-based DAM and DCM (specified in Figure~\ref{fig:architecture}). Out of the $400$ labeled target images, $200$ were employed to select the values of parameters for the DAM by evaluating AUC performance of the formulated target observer on them. The other $200$ labeled target images were employed for the method evaluation. 

The naive method that directly applies the trained source observer to each of the 5 target image sets is referred to as the \textit{Source Observer} (SO). The proposed method that employs adversarial domain adaption will be referred to as the \textit{Source Observer+Domain Adaptation} (SODA). Finally, as a reference, we compute the performance of the CNN-based NO  for the case when there is a large amount of labeled target images.  Of course, the assumption of this work is that such data are not readily available. This method, referred to as the \textit{Target Observer} (TO), employs the the same CNN architectures as the ENN and ONN specified in Figure~\ref{fig:architecture}. In the TO method, semi-online learning strategy~\cite{zhou2019approximating} was employed to directly train 5 observers with labeled target images from each of the five target image domains. In each image domain, $100,000$ background images were used to generate labeled signal-present and signal-absent images by adding noises on the fly during the training process. The same 5 validation sets and 5 testing sets generated in the SODA were employed for validating and testing the 5 trained numerical observers in the TO. The detection performances of the TO will be used as ground truths to evaluate the performance of the proposed method at 5 different levels of domain shifts.

\vspace{-.1cm}
\section{Results}
The ROC curves from SO, SODA, and TO were evaluated on the $200$ pairs of target testing images from each of the five target image domains. The Metz-ROC software was employed to fit the ROC curves~\cite{metz1998rockit}.\begin{wrapfigure}[17]{l}{2.8in}
        \includegraphics[width=\linewidth]{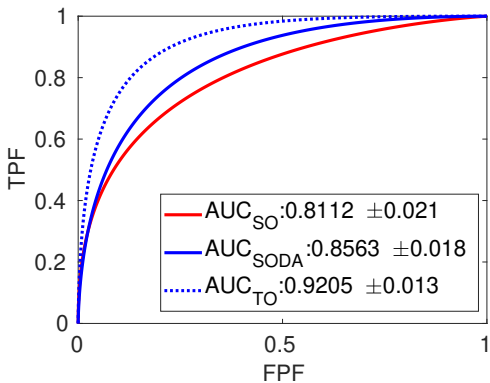}
        \caption{ROC curves corresponding to the SO, SODA and TO in the case where the target images are produced with system blur $w=4.0$.}
        \label{fig:roc}
\end{wrapfigure} As an example, the fitted ROC curves for the three methods computed in the target domain associated with $w=4.0$ is shown in Figure~\ref{fig:roc}. The resulting ROC curve and AUC value of the proposed method were compared to those produced by use of the SO and TO respectively. The AUC values corresponding to the numerical observers trained by use of the SO, SODA, and TO are $0.8112$, $0.8563$, and $0.9205$, respectively. It is observed that, as expected, directly applying a trained source observer to target images provides the worst detection performance due to the domain shift between the source and target domains. By use of a domain adaptation strategy, the proposed method (SODA) can learn a CNN-based numerical observer that shows improved detection performance compared to the SO. The TO shows the best detection performance, however it is trained with a large amount of labeled target images but only a limited amount of \emph{unlabeled} target images were employed in the proposed method (SODA).

Additionally, the detection performance, in terms of AUC, of the proposed SODA as a function of domain shifts are shown in the Table~\ref{tab:domainshift}. 
\begin{table}[!h]
\centering
\begin{tabular}{|c|c|c|c|c|c|}
	\hline
	Target images & $T:2.0$  & $T:3.0$ & $T:4.0$ & $T:5.0$ & $T:6.0$\\
	\hline
	SO & 0.9797	 &0.9225 &0.8112 &0.7009 &0.6147\\
	\hline
	SODA & $\mathbf{0.982}$ & $\mathbf{0.926}$ & $\mathbf{0.8563}$ & $\mathbf{0.7571}$  & $\mathbf{0.6515}$\\
	\hline
	TO & 0.9902	 &0.9657 &0.9205 &0.8499	&0.7942\\
	\hline
\end{tabular}
\caption{The AUC performance evaluated in the 5 domain shift scenarios.}
\label{tab:domainshift}
\end{table}
It can be observed from the table that in all the 5 domain shift cases, the proposed SODA improved the AUC performance compared to the SO, which demonstrate the potential of the proposed method. From the table we can also see that the gap between the AUC performance of proposed method and that of the TO increases with the increase in the level of domain shift and the increase in challenge level of the corresponding adaptation task.

\vspace{-.1cm}
\section{Conclusions}

This study provides a novel method to learn deep learning-based numerical observers operating on experimental data by use of adversarial domain adaptation methods. 
As a proof-of-principle study, a CNN-based numerical observer is learned by use of the proposed strategy for a binary SKE/BKS signal detection task. 
Experimental results demonstrate that the proposed method has the ability to learn deep learning-based numerical observers that operate on unlabeled experimental data in medical imaging. In future, more realistic object models will be employed to investigate the proposed method. Also, other type numeric observers, e.g. linear observers, will be learned to investigate the proposed method.

\acknowledgments 
This research was supported in part by NIH awards EB020604, EB023045, NS102213, EB028652, R01CA233873, R21CA223799, and NSF award DMS1614305.

\bibliography{main-v2} 
\bibliographystyle{spiebib} 

\end{document}